\documentclass[letterpaper, 10 pt, conference]{ieeeconf}  

\IEEEoverridecommandlockouts                              

\usepackage{amsmath, amssymb, bm}
\usepackage{amsmath}
\usepackage{amssymb}
\usepackage{graphicx}
\usepackage[colorlinks=true, allcolors=blue]{hyperref}
\usepackage{booktabs}
\usepackage{todonotes}
\usepackage{tikz}
\usepackage{subcaption}
\usepackage{multirow}
\newcommand{\ImgH}{37mm}
\newcommand\edit[1]{{\color{black}#1}}

\tikzset{inset/.style={
    anchor=north west,
    xshift=0pt, yshift=-0pt,  
}}
\newcommand{\InsetPic}[2][0.50]{
  \includegraphics[width=#1\linewidth]{#2}%
}

\newcommand{\insetover}[6][]{%
  \begin{tikzpicture}
    \node[inner sep=0] (base) {\includegraphics[#1]{#2}};
    \node[anchor=north west, inner sep=0] at ([xshift=#5,yshift=-#5] base.north west)
      {\includegraphics[width=#4,trim=#6,clip]{#3}};
  \end{tikzpicture}%
}

\title{\LARGE \bf
DisFlow: Scene Flow from Distance Field for Object Pose, Velocity Tracking, and Dynamic Object Reconstruction 

}

\author{Lan Wu$^{1,}$$^{2}$, Sheila Sutjipto$^{1}$, Jennifer Wakulicz$^{1}$ and Teresa Vidal-Calleja$^{1}$
\thanks{$^{1}$Robotics Institute, University of Technology Sydney, Ultimo, NSW 2007, Australia. Corresponding author: {\tt\footnotesize Lan.Wu-2@uts.edu.au}}
\thanks{$^{2}$School of Engineering, University of Western Australia, Perth, Australia.}
}

\begin{document}

\maketitle

\begin{abstract}
We present \emph{DisFlow}, a novel framework for online scene flow estimation from distance field that enables \emph{6DoF dynamic object pose estimation}, \emph{motion tracking}, and \emph{surface reconstruction}. The scene is represented by Gaussian Process Implicit Surfaces (GPIS), with surface normals serving as derivative constraints, enabling accurate signed distance computations near the surface and gradient queries with uncertainty. With this representation as a foundation, we compute a scene flow from the distance field that describes how surface points are transported over time in consecutive frames.
Through our flow, we can estimate an object's pose and motion by incrementally registering a new observed point cloud via an elegant closed-form optimisation. Unlike prior methods that operate in the camera or world frame, our approach performs probabilistic fusion directly in the \emph{object frame}, where the object remains geometrically consistent over time. The tight coupling of the DisFlow method in space and time yields dense geometry, surface normals, object pose trajectories, velocities, and uncertainty, all at real-time rates. We evaluate DisFlow on dynamic object sequences and demonstrate that it achieves accurate pose and motion tracking while simultaneously reconstructing high-quality object surfaces. Code publicly available at
\href{https://github.com/LanWu076/disflow_ros2}{https://github.com/LanWu076/disflow\_ros2}
\end{abstract}

\section{Introduction}

Estimating the pose and motion of dynamic objects is a fundamental capability for robotic perception. 
It enables a wide range of downstream tasks, including manipulation, handover, grasping, navigation, and human–robot interaction. 
While classical methods in simultaneous localisation and mapping (SLAM) or reconstruction have achieved remarkable results for static environments, dynamic objects remain challenging. 
The central difficulty lies in jointly reasoning about object motion and surface geometry from partial and noisy observations.

\begin{figure}[t]
  \centering

  \begin{subfigure}[b]{0.48\textwidth}
    \centering
    \includegraphics[width=\linewidth,trim=0pt 100pt 0pt 0pt,clip]{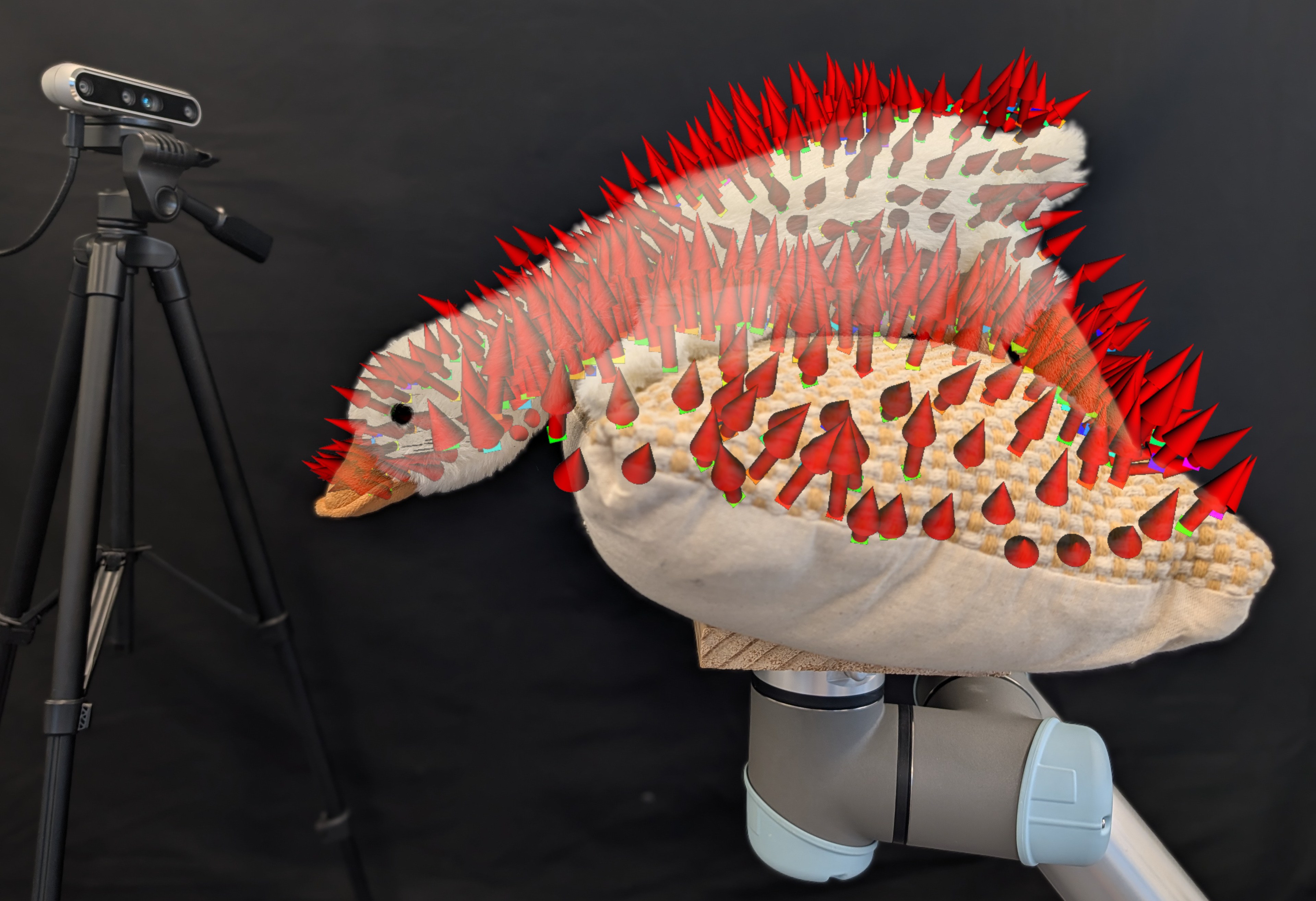}
    \caption{}\label{fig:a}
  \end{subfigure}


  \begin{subfigure}[b]{0.24\textwidth}
    \centering
    \insetover[width=\linewidth,trim=300pt 230pt 150pt 350pt,clip]
              {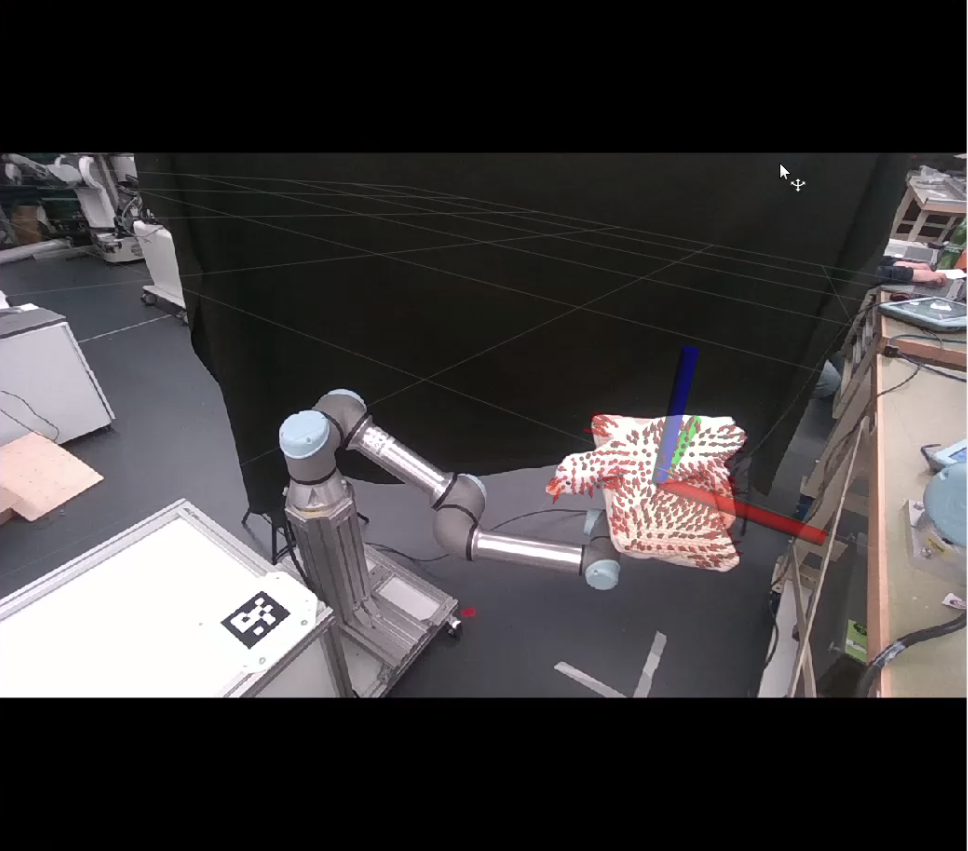}{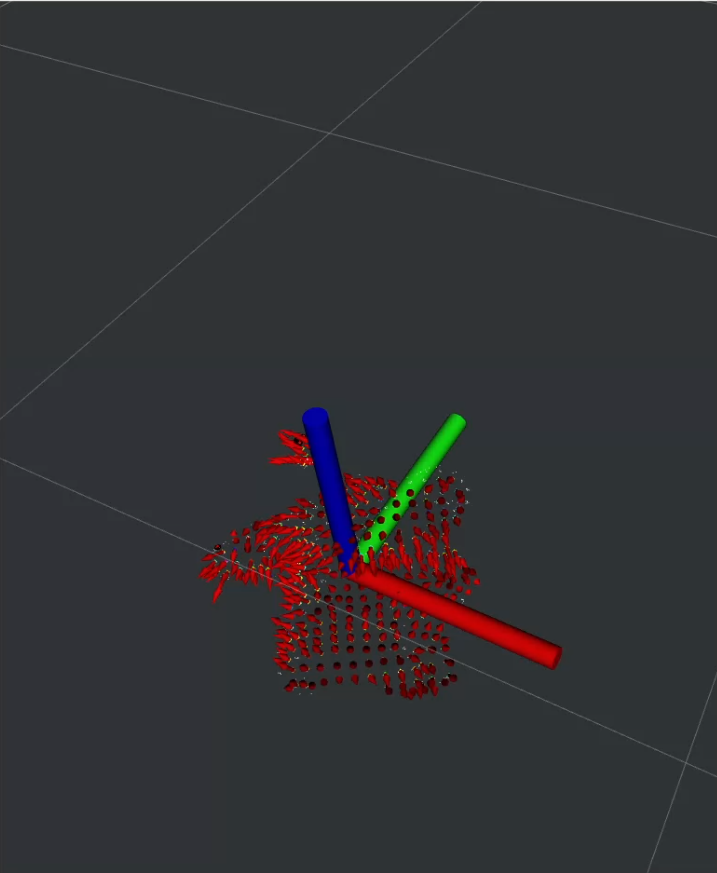}{0.45\linewidth}{2pt}{100pt 100pt 100pt 250pt}
    \caption{}\label{fig:b}
  \end{subfigure}
  \begin{subfigure}[b]{0.24\textwidth}
    \centering
    \insetover[width=\linewidth,trim=300pt 230pt 150pt 350pt,clip]
              {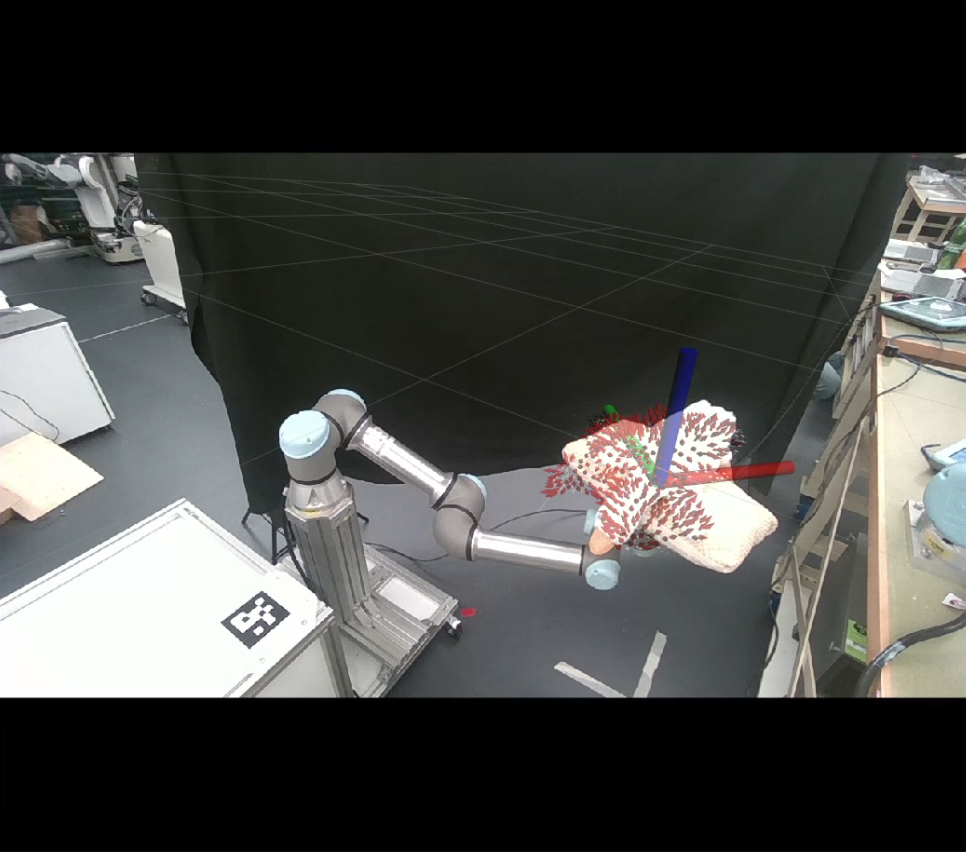}{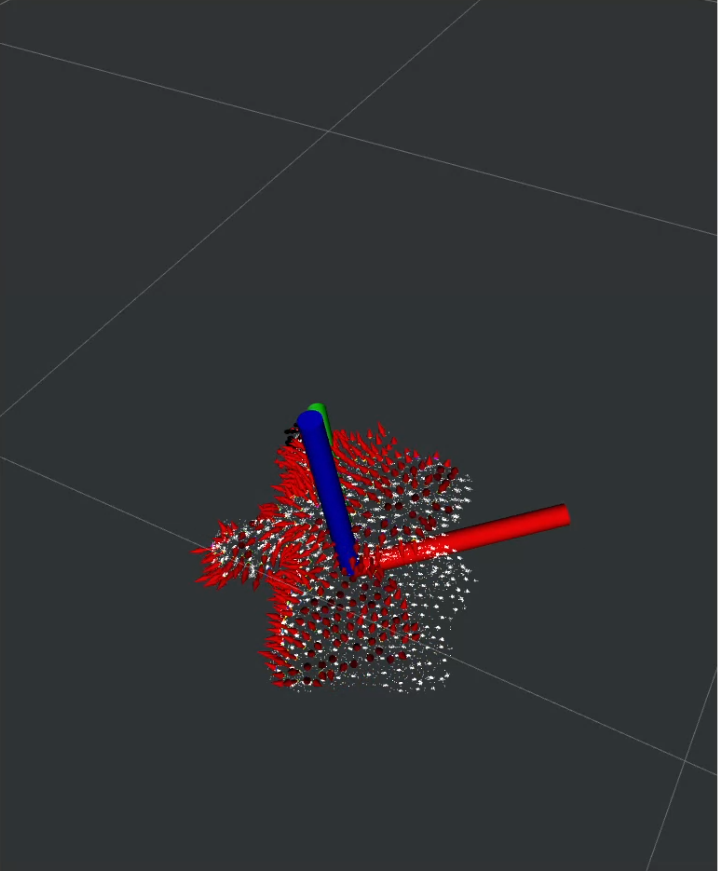}{0.45\linewidth}{2pt}{100pt 100pt 100pt 250pt}
    \caption{}\label{fig:c}
  \end{subfigure}


  \begin{subfigure}[b]{0.24\textwidth}
    \centering
    \insetover[width=\linewidth,trim=300pt 230pt 150pt 350pt,clip]
              {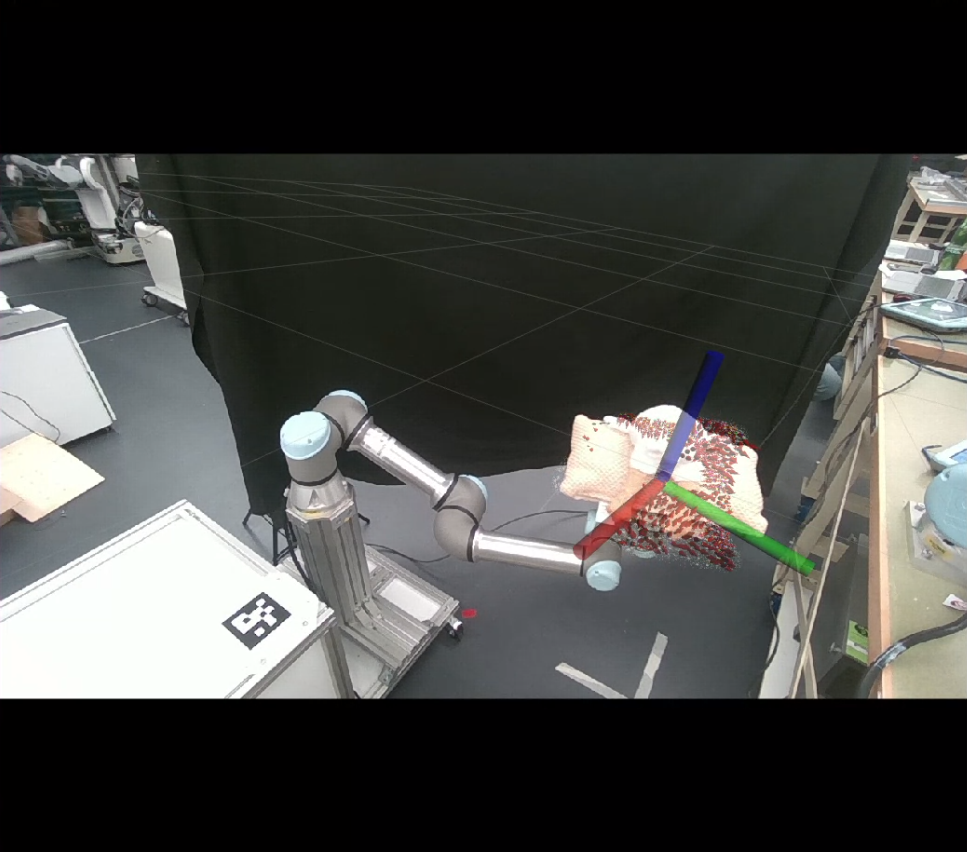}{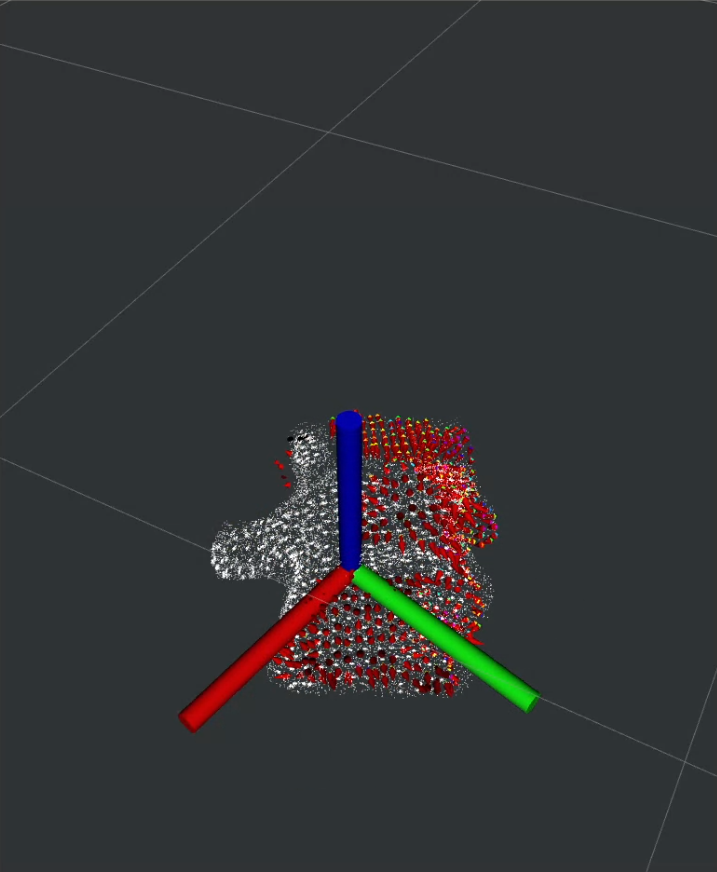}{0.45\linewidth}{2pt}{100pt 100pt 100pt 250pt}
    \caption{}\label{fig:d}
  \end{subfigure}
  \begin{subfigure}[b]{0.24\textwidth}
    \centering
    \insetover[width=\linewidth,trim=300pt 230pt 150pt 350pt,clip]
              {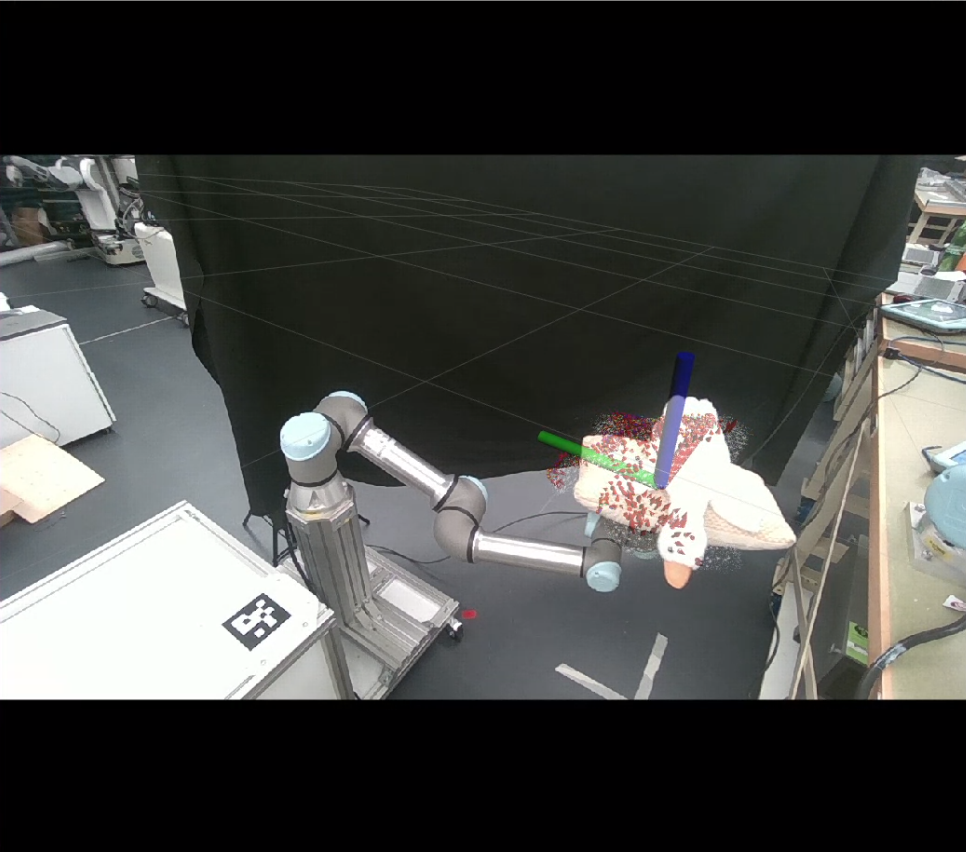}{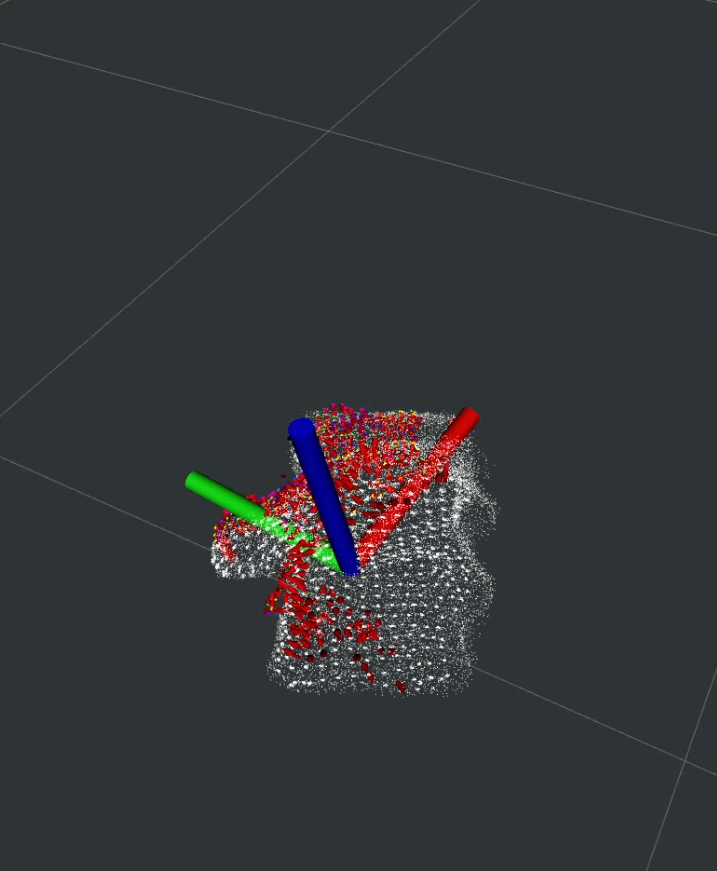}{0.45\linewidth}{2pt}{100pt 100pt 100pt 250pt}
    \caption{}\label{fig:e}
  \end{subfigure}


  \caption{The output of \textit{DisFlow} framework and the experimental setup (a). Subfigures show the fixed object reconstruction in the object frame on the left, and the same reconstruction overlaid with the rotating RGB stream on the right (b)-(e). Refer to the supplementary video for details.}

  \label{fig:combined}
  \vspace{-3ex}
\end{figure}




Recent works have explored real-time object tracking using either filtering based on optical flow~\cite{ROFT} or learning based pipelines~\cite{Labbe2020CosyPose}. However, such methods focus primarily on pose tracking and estimating dense flow, but not on maintaining a dense surface reconstruction of the object itself. 
Therefore, they are limited to tasks that require detailed surface information, such as contact reasoning or manipulation. In addition, learning-based pipelines rely on large-scale training data and are typically trained on specific objects, which limits their generalisation to unseen categories.


In contrast, dense reconstruction approaches such as TSDF fusion~\cite{Curless1996Volumetric} or volumetric mapping~\cite{millane2024nvblox} build explicit object surfaces but are typically designed for static scenes.
Moreover, conventional volumetric fusion lacks a probabilistic representation and does not provide uncertainty, which is important for safety and planning.

In this work, we present \emph{DisFlow}, a novel framework for online scene flow estimation from distance field, enabling 6DoF object pose and velocity estimation with online surface reconstruction.
Our method relies on an object-centric reference frame to perform object pose tracking and fusion in a temporally consistent manner (see Fig.~\ref{fig:combined}.
We represent the object with Gaussian Process Implicit Surfaces (GPIS)~\cite{GPbook,paper:GPImplSurf}, trained with both surface points and normals as derivative constraints. The distance field from GPIS is accurate close to the surface, given a high frame rate.
This enables accurate signed distance and gradient queries, along with uncertainty estimates. Grounded on this representation, we compute the scene flow~\cite{vedula1999three} based on the distance field that describes how surface points are transported over time as the object moves. This notion of flow provides the conceptual link between pose tracking and surface reconstruction. Through our flow, we can estimate an object’s pose and motion by incrementally registering a new observed point cloud via an elegant closed-form optimisation.
The estimated pose then projects new observations into the object frame, incrementally updating local GPs stored in an octree structure. 
This tight integration yields real-time outputs of object pose, velocity, dense surface reconstruction, and associated uncertainty.

The key contributions of this paper are:
\begin{itemize}
    \item We propose \textbf{DisFlow}, a framework that leverages instantaneous distance fields to recover the scene flow, enabling 6DoF and linear and angular velocities with the probabilistic surface reconstruction in real-time.
    \item We introduce an \textbf{object-centric strategy} that naturally handles dynamic objects, ensuring consistent geometry.
    \item We demonstrate that DisFlow simultaneously delivers \textbf{real-time pose trajectories, object velocities, and dense surface reconstructions}, outperforming tracking-only or reconstruction-only baselines, and providing a foundation for downstream applications.
\end{itemize}

\section{Related Work}

\subsection{Object Pose and Motion Tracking}
Estimating dynamic objects, including 6DoF pose and velocity, has been extensively studied in robotics and computer vision. 
Classical approaches such as Iterative Closest Point (ICP) and its variants~\cite{Besl1992ICP,Segal2009GeneralizedICP} perform frame-to-frame alignment of point clouds, but remain sensitive to initialisation and partial views, limiting their robustness in dynamic scenarios. 

Flow-based formulations have long been explored as an alternative. 
Normal flow along image gradients provides a directly observable constraint~\cite{horn1981determining}, while scene flow generalises this idea to RGB-D data to estimate dense 3D motion fields~\cite{vedula1999three}. 
Recent works leverage dense flow for pose refinement, e.g., DeepIM~\cite{li2018deepim} aligns rendered and observed images, and ROFT~\cite{ROFT} formulates pose tracking as robust optical flow alignment to compensate for network latency. 
These methods highlight the utility of flow cues, but typically depend on strong learning priors or hand-crafted constraints, and seldom provide principled uncertainty estimates. 

Beyond flow-based methods, learning-driven systems such as PoseCNN~\cite{xiang2017posecnn}, PVNet~\cite{Peng2019PVNet}, and CosyPose~\cite{Labbe2020CosyPose} have demonstrated improved robustness across object categories. 
Other pipelines, such as Co-Fusion~\cite{Runz2017CoFusion} address dynamic objects by fusing depth in the world frame. 
Despite these advances, most methods either prioritise tracking accuracy without maintaining dense and temporally consistent surface reconstructions or rely on heavy GPU computation, limiting their deployability in real-time robotic settings.

\subsection{Volumetric Reconstruction}
Dense 3D reconstruction has been widely addressed using volumetric signed distance fields, starting from the work~\cite{Curless1996Volumetric}. 
KinectFusion~\cite{kinectfusion} demonstrated real-time surface reconstruction by fusing depth images into a global TSDF volume, inspiring numerous extensions such as Nvblox~\cite{millane2024nvblox}. 
While these systems achieve high-quality reconstructions in static scenes, handling dynamics remains challenging. 
Approaches such as Co-Fusion~\cite{Runz2017CoFusion} incorporate non-rigid alignment or segmentation to deal with moving objects, yet they still rely on rule-based filtering to discard inconsistent observations.
Moreover, TSDF-based fusion does not provide uncertainty estimates, which are critical for safe interaction in robotics.

\subsection{Probabilistic Implicit Representations}
To overcome the limitations of discrete volumetric grids, probabilistic implicit representations have been explored. 
Gaussian Process Implicit Surfaces (GPIS)~\cite{paper:GPImplSurf} provide a continuous and differentiable model of object surfaces, capable of jointly predicting signed distances and gradients. 
Extensions incorporating normals as derivative constraints~\cite{bhoram_online_2019} improve surface fidelity and orientation consistency. 
Such models naturally provide uncertainty estimates, which have been exploited for active perception and safety-aware planning. 
However, prior GPIS-based works typically focus on static scenes or offline reconstruction, and have not been demonstrated for real-time dynamic tracking.

In summary, tracking-oriented methods deliver real-time 6D poses and velocities but lack dense geometry, while reconstruction-oriented methods provide detailed surfaces but struggle with dynamic objects and uncertainty. 
Probabilistic implicit models offer a promising alternative but have not been applied to real-time dynamic object tracking and reconstruction. 
Our work addresses this gap by introducing \emph{DisFlow}, which exploits the scene flow from the distance field to estimate 6D object pose and velocity tracking with online surface reconstruction in the object frame, leveraging Gaussian Process Implicit Surfaces for probabilistic and uncertainty-aware modelling.

\section{Notation and Problem Definition}

Let us define the \textbf{Camera frame} $\mathcal{F}_c$ attached to the optical centre of the RGB-D sensor. 
At each time step $k$, the sensor provides a set of points:
$\{\mathbf{x}_{k,j}^c \in \mathbb{R}^3 \mid j=1,\dots,M\}$, where $\mathbf{x}_{k,j}^c$ denotes the 3D position of the $j$-th point of frame $k$ expressed in the camera frame. 

The \textbf{World frame} $\mathcal{F}_w$ is a global reference frame typically used for mapping. We are interested in the relative motion between a moving object and a camera. 
Depending on whether the camera is moving, $\mathcal{F}_c$ may coincide with $\mathcal{F}_w$ 
(\emph{i.e.}, $T^k_{wc}=I$ $\forall k$) or vary over time. Let us also define the \textbf{Object frame} $\mathcal{F}_o$, which is rigidly attached to the object of interest. 
At time step $k$, the transformation from the camera frame to the object frame is denoted:
\begin{equation}
T_{oc}^k =
\begin{bmatrix}
R^k_{oc} & \mathbf{t}^k_{oc} \\
\mathbf{0}^\top & 1
\end{bmatrix} \in SE(3),
\end{equation}
where $R^k_{oc} \in SO(3)$ is the rotation and $\mathbf{t}^k_{oc} \in \mathbb{R}^3$ is the translation. 
This transformation maps a point $\mathbf{x}_{k,j}^c$ in the camera frame into the object frame as
\begin{equation}
\mathbf{x}_{k,j}^o = T_{oc}^k \, \mathbf{x}_{k,j}^c.
\end{equation}



Given the sequence of 3D points produced by the RGB-D camera, our goal is to recover the dense flow induced by the relative motion between the camera and the object. 
Moreover, from this flow, we aim to estimate the instantaneous object pose $T^k_{oc}$ and its linear velocity $\mathbf{v}$ and angular velocity $\boldsymbol{\omega}$, 
while at the same time fusing $\mathbf{x}_{k,j}^o$ into a temporally consistent implicit surface representation parameterised as a signed distance field $d$.

\section{Implicit Surface Representation}

To represent the dynamic object geometry, we employ \emph{Gaussian Process Implicit Surfaces} (GPIS)~\cite{paper:GPImplSurf,wu_faithful_2021,wu2022mop,gentil_accurate_2023}. Unlike discrete volumetric grids such as TSDFs or occupancy maps, a GPIS provides a continuous, probabilistic representation of the surface, offering both signed distance predictions around the surface and uncertainty estimates. 

\paragraph{Implicit Surface Formulation} 
Given a set of transformed surface points $\{\mathbf{x}_{j}^o\}_{j=1}^M$ from the object frame, let us define an implicit function $d: \mathbb{R}^3 \rightarrow \mathbb{R}$ to model the distance around the surface such that the surface is the zero level set:
\begin{equation}
\mathcal{S} = \{\mathbf{x}_{j}^o \in \mathbb{R}^3 \mid d(\mathbf{x}_{j}^o) = 0\}.
\end{equation}
The sign of $d$ indicates whether the point lies inside or outside the object. In practice, $d$ is modelled using a Gaussian Process (GP) prior with mean $\mu(\mathbf{x})$ and kernel $k(\mathbf{x}, \mathbf{x}')$, yielding the distribution
\begin{equation}
d(\mathbf{x}) \sim \mathcal{GP}\big(\mu(\mathbf{x}), \; k(\mathbf{x},\mathbf{x}')\big).
\end{equation}

\paragraph{Training Constraints} 
Two types of constraints:
\begin{enumerate}
    \item \textbf{Surface constraints:} Every single observed surface point should lie on the zero level set:
    $d(\mathbf{x}_{j}^o) = 0$.
    \item \textbf{Normal constraints:} The gradient at ${d}_i$ should align with the observed normal $\mathbf{n}_i$, i.e.
    $\nabla d(\mathbf{x}_{j}^o) \approx \mathbf{n}_i.$
    In practice, normals are estimated from local neighbourhoods and flipped consistently with the camera viewpoint to ensure an outward orientation.
\end{enumerate}
Incorporating gradient observations effectively constrains the first-order derivatives of the kernel function, improving the stability and sharpness of the reconstructed surface.

\paragraph{Kernel Design}
We use a Matérn $\tfrac{3}{2}$ kernel, which provides a good balance between smoothness and local adaptability~\cite{bhoram_online_2019}. 

\paragraph{Prediction}
For a query point $\mathbf{x}_*$, the posterior distribution of the signed distance and its gradient is
\begin{equation}
\begin{bmatrix} d(\mathbf{x}_*) \\ \nabla d(\mathbf{x}_*) \end{bmatrix}
\sim \mathcal{N}\!\left(
\begin{bmatrix} d(\mathbf{x}_*) \\ \nabla d(\mathbf{x}_*) \end{bmatrix},
\begin{bmatrix} V[d_*] & V[d_*,\nabla d_*] \\ V[\nabla d_*,d_*] & V[\nabla d_*] \end{bmatrix}
\right),
\end{equation}
where the $V$ is the variance. The mean and covariance are:
\begin{equation}
\begin{bmatrix} d(\mathbf{x}_*) \\ \nabla d(\mathbf{x}_*) \end{bmatrix}
= \tilde k_*^\top \big(\tilde K+\sigma^2 I\big)^{-1}\mathbf{y},
\label{eq:gpis_query}
\end{equation}
\begin{equation}
\begin{bmatrix} V[d_*] & V[d_*,\nabla d_*] \\ V[\nabla d_*,d_*] & V[\nabla d_*] \end{bmatrix}
= \tilde k(\mathbf{x}_*,\mathbf{x}_*) - \tilde k_*^\top \big(\tilde K+\sigma^2 I\big)^{-1}\tilde k_*.
\end{equation}
Here $\tilde{k}(\mathbf{x}, \mathbf{x}')$ is the joint covariance function of the signed distance and its derivatives~\cite{paper:GeomPrior,GPbook}.

\paragraph{Scalability via Octree}
A naive full GP model suffers from cubic complexity in the number of training samples. To enable real-time operation, we adopt a spatial partitioning strategy, similar to the work in~\cite{bhoram_online_2019,ali2024interactive}. The 3D domain is divided by an octree, and each cluster node maintains a \emph{local GP} trained from nearby samples. At runtime, a query point is directed to the relevant local GP, ensuring efficient prediction. This design offers several advantages. Each GP models a compact spatial region, improving numerical stability. As new frames arrive, only the affected local GPs need to be retrained, making online fusion feasible for incremental updates. The octree naturally supports hierarchical queries, allowing for coarse-to-fine resolution control.

This representation provides a continuous, differentiable, and probabilistic implicit model of the object surface. Compared to voxel-based TSDFs, GPIS can predict geometry at arbitrary resolutions and directly supply surface normals from $\nabla d$. Compared to purely tracking-based systems, it offers a dense surface model that is temporally consistent in the object frame, serving as the foundation for both reconstruction and motion estimation.

\section{DisFlow Framework}

To recover the dense flow that describes the motion of a moving object with respect to a depth camera, we formulate the problem as the temporal change of the distance field $\frac{\partial d(\mathbf{x},t)}{\partial t}$ between consecutive frames $k-1$ and $k$ described by the transport equation. Given the GPIS representation, we compute the scene flow by querying distance and gradients from one frame to the other, we called this flow \emph{DisFlow}. Our DisFlow describes how surface points are transported over time as the object moves. 

\subsection{Object-Centric Frame}
Dynamic object reconstruction and tracking involve reasoning over multiple coordinate frames. In this work, the object geometry is represented and incrementally fused in the object frame $T_{oc}$. As a result, the object surface remains consistent across time even when the object moves.
Traditional approaches often reconstruct in the camera or world frame. This requires explicit handling of dynamic points, since the object moves relative to those frames. Such strategies rely on temporal consistency checks to eliminate inconsistent points, which increases complexity and often introduces failure cases in cluttered scenes. In contrast, by maintaining the map directly in the object frame, the object surface is inherently static and geometrically consistent. This eliminates the need for dynamic point removal, simplifies the pipeline, and allows our method to jointly deliver accurate geometry, motion, and pose estimation.

\subsection{Distance Flow Formulation}

Formally, let us define the DisFlow
as the temporal change of the object pose, induced by the motion of the object, the camera or both. 
Let $d(\mathbf{x}, t)$ denote the signed distance function of the object surface at time $t$, considering that the distance field not only changes with location but it might also change with time due to the motion of the scene wrt the camera. 
Based on the distance field that varies over time, the transport equation over distance is given by:
\begin{equation}
\frac{\partial d(\mathbf{x}, t)}{\partial t} + v(\mathbf{x}, t) \nabla d(\mathbf{x}, t) = 0,
\end{equation}
where $v(\mathbf{x},t)$ is the scene flow at $\mathbf{x}$.

This expression describes how the surface points are transported consistently with the object’s rigid motion by the scene flow. Given the GPIS representation, both $d(\mathbf{x})$ and $\nabla d(\mathbf{x})$ are available by simply querying our representation through the analytical expression of \eqref{eq:gpis_query}. 
By querying them at a high frame rate, we obtain a continuous flow field that describes how points move with the object. 
Conceptually, this flow defines implicit point-to-point correspondences between successive observations. 
\edit{We use this implicit knowledge in our optimisation through distance residuals, 
where $\nabla d(\mathbf{x})$ acts as the correspondence direction.}

\subsection{Optimisation Objective}
At time step $k$, let $\{\mathbf{x}_j^c\}_{j=1}^M$ be the input point cloud expressed in the camera frame $\mathcal{F}_c$. 
Under a candidate object pose $T_{oc}^k = (R^k_{oc}, \mathbf{t}^k_{oc}) \in SE(3)$, the points are transformed into the object frame as $\mathbf{x}_j^o$.
For each transformed point $\mathbf{x}_j^o$, the GP distance provides both the signed distance $d(\mathbf{x}_j^o)$ and its gradient $\nabla d(\mathbf{x}_j^o)$. 
Ideally, if $T_{oc}^k$ is correct, transformed points should lie on the implicit surface, i.e.
\[
d(\mathbf{x}_j^o) \approx 0.
\]

We thus define the pose estimation objective function as the sum of squared signed distances:
\[
E = \sum_{j=1}^M d(R^k_{oc} \mathbf{x}_j^c + \mathbf{t}^k_{oc})^2.
\]
Minimising $E$ aligns the observed point cloud with the GPIS surface.
The flow constraints encoded by $d$ and $\nabla d$ guide each point toward its surface location. 

To tackle this optimisation problem, we linearise around the current estimated pose to solve in closed form.

\subsection{Linearisation and Jacobian}
Let $\xi \in \mathfrak{se}(3)$ be a small twist increment, with translation $\delta\mathbf{t}$ and rotation vector $\delta\boldsymbol{\theta}$.
The first-order approximation of the residual for point $\mathbf{x}_j^o$ is
\[
d(\mathbf{x} + \delta \mathbf{x}) \approx d(\mathbf{x}) + \nabla d(\mathbf{x})^\top \delta \mathbf{x}.
\]
Since the perturbation in $\mathbf{x}$ induced by $\xi$ is
\[
\delta \mathbf{x} = \delta \mathbf{t} + \delta\boldsymbol{\theta} \times \mathbf{x}.
\]
The Jacobian of the residual is
\[
J(\mathbf{x}) =
\begin{bmatrix}
\nabla d(\mathbf{x})^\top & (\mathbf{x} \times \nabla d(\mathbf{x}))^\top
\end{bmatrix}.
\]
Stacking across all points yields the normal equations
\[
H = \sum_{j=1}^M J(\mathbf{x}_j)^\top J(\mathbf{x}_j), \quad
b = \sum_{j=1}^M J(\mathbf{x}_j)^\top d(\mathbf{x}_j).
\]
We solve the linear system
\[
H \xi = -b
\]
for the twist $\xi$, which is applied via the exponential map
\[
T \leftarrow \exp(\hat{\xi}) T,
\]
where $\hat{\xi} \in \mathfrak{se}(3)$ is the matrix representation of the twist.


\subsection{Practical Considerations}\label{sec:method:practical}
While the above formulation is purely geometric, in practice, it may be sensitive to noise or partial observations. 
Where motion is smooth, we can improve robustness by introducing a temporal regularisation term that enforces smoothness between consecutive frames. 
Let $v_{k-1}$ be the twist estimated at the previous frame. 
We propagate this into the current frame using the adjoint transformation:
\[
v_{k-1}^\prime = \mathrm{Ad}(T_{o_k o_{k-1}})\, v_{k-1}.
\]
Then, a quadratic penalty
\[
E_{\text{reg}}(\xi) = \|\xi - v_{k-1}^\prime\|_W^2
\]
is added to the objective, where $W$ is a diagonal weight matrix that can independently penalise translation and rotation. 
This encourages smooth motion trajectories and lowers frame-to-frame drift. Furthermore, to ensure stable optimisation, we adopt the following strategies:
\begin{itemize}
    \item \textbf{Residual truncation:} points with $|d(\mathbf{x})|$ larger than a threshold are ignored, since far-away points provide little information and can destabilise optimisation.
    \item \textbf{Gradient weighting:} residuals are weighted by the norm of $\nabla d(\mathbf{x})$, down-weighting uncertain directions in flat regions.
    \item \textbf{Damping:} Levenberg--Marquardt style damping is applied to $H$, improving numerical conditioning and preventing overshooting.
\end{itemize}

DisFlow yields the updated camera-to-object pose $T_{oc}^k$ and the twist $v_k$, which together define the object’s linear and angular velocities in 3D. 
The optimisation finds the rigid-body motion most consistent with the flow constraints implied by $(d,\nabla d)$. 
The resulting DisFlow not only provides accurate pose and velocity estimates, but also establishes the transport needed for fusing new observations into a temporally consistent object-centric model.

\subsection{Object-Frame Fusion}

After estimating the relative pose $T_{oc}^k$ at frame $k$, the observed point cloud can be consistently fused into the object model. 
The transformed set $\{\mathbf{x}_j^o\}$ represents surface samples that are stationary with respect to $\mathcal{F}_o$.

The object domain is partitioned into an octree, where each cluster node maintains a local Gaussian Process trained from the points within its region. 
When new samples $\mathcal{D}_{\text{new}} = \{\mathbf{x}_j^o, \mathbf{n}_j^o\}$ arrive, they are added to the corresponding cluster
\[
\mathcal{D}_{\text{cluster}} \leftarrow \mathcal{D}_{\text{cluster}} \cup \mathcal{D}_{\text{new}},
\]
and the local GP is retrained to refine the signed distance and gradient predictions. 
Since only the clusters intersecting the new observations are updated, the procedure is efficient and supports incremental online operation.

In the object frame, the surface is by definition temporally consistent, so fusion is straightforward and robust even under fast motions. 
As more points are added, the local GPs become more accurate in well-observed areas, while still keeping high uncertainty in regions that have not been seen much.
The result is an incrementally refined probabilistic surface model of the object, which simultaneously supports distance flow, pose tracking, dense reconstruction, and downstream robotic tasks.

\subsection{Real-Time Outputs}

Our framework produces several outputs in real-time. 
The fusion yields a continuous and probabilistic implicit surface, providing dense geometry of the object together with normals and gradients obtained directly from the GP predictions. 
In parallel, our DisFlow delivers both the object pose $T_{oc}^k$ and the velocity $v_k$. 
Finally, each GP query also returns an estimate of uncertainty via the posterior variance, which reflects reconstruction confidence and can be exploited in planning, grasping, or human–robot interaction. 
All outputs are available online through standard ROS topics and services, enabling integration into robotic systems.


\section{Evaluation}

We evaluate the proposed \textit{DisFlow} framework using both quantitative and qualitative experiments. The quantitative evaluation is performed on the Fast-YCB dataset, which provides ground-truth for object pose, velocity, and point cloud. This enables us to evaluate three key aspects and adopt the results from the ROFT paper~\cite{ROFT}. First, object pose tracking is evaluated with ADD-AUC and RMSE metrics, and compared against DOPE~\cite{tremblay2018dope}, PoseRBPF\cite{deng2021poserbpf}, TrackNet\cite{huang2019tracknet}, and ROFT~\cite{ROFT}. Second, velocity tracking is evaluated using RMSE for linear and angular velocity errors, and is compared directly with ROFT. Third, surface reconstruction is evaluated with Chamfer distance and normal consistency, and compared against TSDF-based fusion.  

In addition to the quantitative evaluation, we also conduct qualitative experiments on real sequences. The first setup moves an object around in the scene and brings it back to the starting position. The second setup records a person rotating a full circle and returning to the starting viewpoint. These two setups test the consistency and robustness of our framework to deal with dynamic object tracking.

\subsection{Object Pose Tracking}

\begin{table}[t]
\centering
\caption{Results on the Fast-YCB dataset for 003\_cracker\_box.}
\label{tab:pose_results_003}
\begin{tabular}{lccc}
\toprule
Method & ADD-AUC (\%) & RMSE $e_t$ (cm) & RMSE $e_a$ (deg) \\
\midrule
DOPE        & 54.92 & 5.1 & 28.33 \\
ROFT        & 78.50 & 2.5 & 7.55 \\
PoseRBPF    & 68.94 & 2.6 & 38.46 \\
TrackNet    & 63.02 & 7.9 & 31.73 \\
DisFlow (ours) & \textbf{92.44} & \textbf{0.790} & \textbf{3.519} \\
\bottomrule
\end{tabular}
\end{table}

\begin{table}[t]
\centering
\caption{Results on the Fast-YCB dataset for 006\_mustard\_bottle.}
\label{tab:pose_results_006}
\begin{tabular}{lccc}
\toprule
Method & ADD-AUC (\%) & RMSE $e_t$ (cm) & RMSE $e_a$ (deg) \\
\midrule
DOPE        & 57.20 & 8.6 & 36.13 \\
ROFT        & 73.10 & 3.1 & 13.29 \\
PoseRBPF    & 82.92 & 2.0 & 18.42 \\
TrackNet    & 74.83 & 11.9 & 35.48 \\
DisFlow (ours) & \textbf{94.83} & \textbf{0.498} & \textbf{5.038} \\
\bottomrule
\end{tabular}
\end{table}

We first evaluate the accuracy of object pose tracking on the Fast-YCB dataset. 
Following the protocol in~\cite{ROFT, xiang2017posecnn}, we report three metrics: the ADD-AUC, the RMSE of translation error $e_t$, and the RMSE of angular error $e_a$. These metrics capture the geometric alignment, translational stability, and orientation consistency of the estimated trajectories. For each frame $k$, the Average Distance (ADD) error is computed as
\begin{equation}
\text{ADD}_k = \frac{1}{|\mathcal{M}|} \sum_{x \in \mathcal{M}} \left\| \hat{T}_k x - T_k x \right\|,
\end{equation}
where $\mathcal{M}$ is the set of 3D model points, $T_k \in SE(3)$ is the ground-truth pose, and $\hat{T}_k$ is the estimated pose.  

The accuracy for a given distance threshold $d$ is
\begin{equation}
\text{acc}(d) = \frac{1}{N} \sum_{k=1}^N 
\begin{cases}
1, & \text{if } \text{ADD}_k < d, \\
0, & \text{otherwise}.
\end{cases}
\end{equation}
$N$ is the number of frames.  
The ADD-AUC score is then obtained by numerical integration up to $\tau = 0.10$\,m:
\begin{equation}
\text{ADD-AUC} = \frac{1}{\tau} \int_0^\tau \text{acc}(d)\,\mathrm{d}d.
\end{equation}


We compare \textit{DisFlow} against four baselines: DOPE~\cite{tremblay2018dope}, ROFT~\cite{ROFT}, PoseRBPF~\cite{deng2021poserbpf}, and TrackNet~\cite{huang2019tracknet}. DOPE is a per-frame deep network that provides absolute poses. PoseRBPF propagates pose hypotheses over time, but it is computationally expensive and prone to track loss. se(3)-TrackNet is an end-to-end network trained on synthetic data to regress inter-frame transformations. ROFT requires DOPE predictions as inputs. It has an optical-flow-aided Kalman filter to synchronise delayed outputs and provide both pose and velocity estimates. Our proposed DisFlow requires only point clouds as input and fuses observations directly in the object frame using Gaussian Process Implicit Surfaces, yielding temporally consistent tracking and dense geometry.

Table~\ref{tab:pose_results_003} shows results for the \texttt{003\_cracker\_box}, a box-shaped object with sharp edges and strong geometric constraints. 
As expected, DOPE alone suffers from low ADD-AUC accuracy, while PoseRBPF and TrackNet improve the ADD-AUC accuracy but remain unstable in translation and angular estimates. ROFT improves over these baselines, reaching an ADD-AUC of 78.50\% and angular error of 7.55$^\circ$. Our DisFlow further improves the performance substantially, achieving an ADD-AUC of 92.44\%, a translation error of 0.790 cm, and a low angular error of 3.519$^\circ$. This demonstrates that our method can reliably exploit surface geometry to maintain accurate tracking under fast motions.

Table~\ref{tab:pose_results_006} presents results for the \texttt{006\_mustard\_bottle}, which poses a much harder challenge due to its smooth cylindrical shape, low structure, and near symmetry. In this case, DOPE and TrackNet again suffer from both translation and angular drift, while PoseRBPF achieves lower translation errors but inconsistent rotations. ROFT improves angular stability, but its ADD-AUC remains below 75\%. DisFlow can overcome these challenges and achieve the best results: an ADD-AUC of 94.83\%, a translation error of only 0.498~cm, and an angular error of 5.038$^\circ$. This demonstrates that our framework is able to maintain robust tracking performance even for objects with weak geometric constraints, outperforming all compared baselines by a clear margin.

\subsection{Velocity Tracking}

Accurate velocity tracking is essential for dynamic interaction, as it enables the prediction of future object states and facilitates safe and responsive robot control. While many existing approaches, such as DOPE, PoseRBPF, and TrackNet, only provide per-frame pose estimates, they do not output object velocities. Among the prior methods, only ROFT~\cite{ROFT} explicitly estimates 6D velocities, making it the primary baseline for comparison. We therefore focus on a direct quantitative evaluation against ROFT using the Fast-YCB dataset, which provides ground-truth linear and angular velocities for every frame.

Table~\ref{tab:vel_results} reports results for the \texttt{003\_cracker\_box}. ROFT achieves a linear velocity RMSE of 5.237~cm/s and an angular velocity RMSE of 18.250$^\circ$/s. In contrast, DisFlow achieves 2.311~cm/s and 7.109$^\circ$/s, corresponding to more than a two-fold reduction in both translation and rotation errors. This demonstrates that our object-frame formulation, which fuses observations probabilistically into GPs, provides a stable estimate of object dynamics under rapid motion.

Results for the \texttt{006\_mustard\_bottle} are shown in Table~\ref{tab:vel_results}. This object is more challenging due to its smooth cylindrical geometry and partial symmetries, which often cause conventional trackers to drift or produce unstable velocity estimates. Nevertheless, DisFlow again improves over ROFT. The linear velocity error is reduced from 7.227 to 1.988~cm/s, and the angular error from 26.553$^\circ$/s to 20.343$^\circ$/s. These results confirm that our method maintains robustness and temporal consistency even in the presence of near-symmetric shapes.

In summary, DisFlow consistently outperforms ROFT on both translational and rotational velocity estimation. By coupling velocity tracking directly with the probabilistic representation and distance flow, our approach delivers velocity predictions that complement pose tracking and enable real-time use in manipulation and control.



\begin{table}[t]
\centering
\caption{Velocity tracking results on the Fast-YCB dataset.}
\label{tab:vel_results}
\begin{tabular}{l lcc}
\toprule
Object & Method & RMSE $e_v$ (cm/s) & RMSE $e_\omega$ (deg/s) \\
\midrule
\multirow{2}{*}{003\_cracker} 
    & ROFT            & 5.237 & 18.250 \\
    & DisFlow  & \textbf{2.311} & \textbf{7.109} \\
\midrule
\multirow{2}{*}{006\_mustard} 
    & ROFT            & 7.227 & 26.553 \\
    & DisFlow  & \textbf{1.988} & \textbf{20.343} \\
\bottomrule
\end{tabular}
\end{table}

\subsection{Surface Reconstruction}

Finally, we evaluate the quality of dense surface reconstruction. This is important for robotic interaction, as downstream tasks such as grasp planning and collision checking require not only pose and velocity but also a reliable 3D mesh representation. The Fast-YCB dataset provides a ground-truth point cloud, which allows both quantitative and qualitative comparison.

\begin{figure*}[t]
  \centering
  \subfloat[GT\label{mesh:gt}]{
    \includegraphics[width=2cm,keepaspectratio]{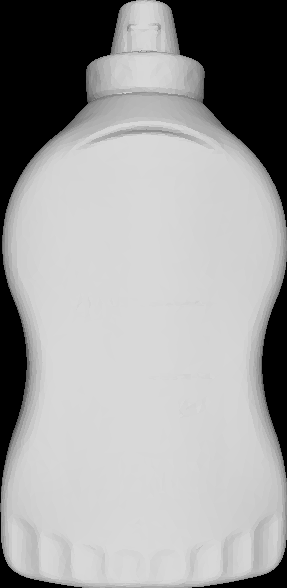}}
  \hspace{1mm}
  \subfloat[TSDF\label{mesh:tsdf}]{
    \includegraphics[width=1.95cm,keepaspectratio]{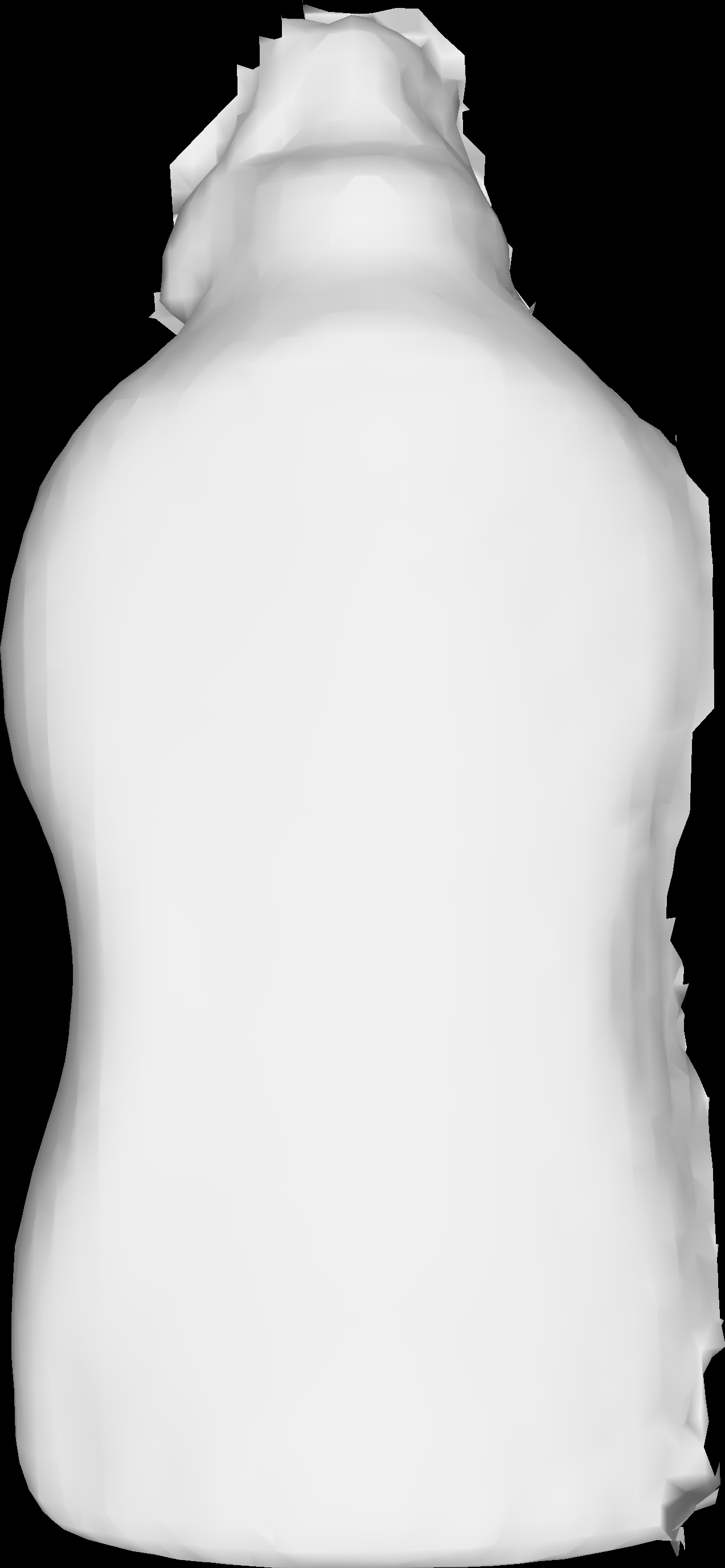}}
  \hspace{1mm}
  \subfloat[DisFlow\label{mesh:ours}]{
    \includegraphics[width=2.05cm,keepaspectratio]{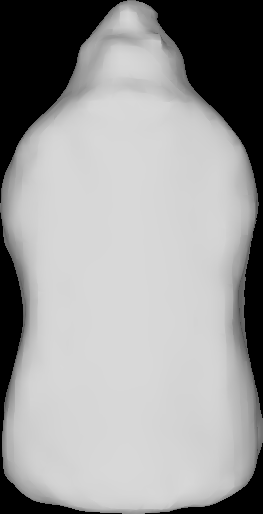}}
  \hspace{1mm}
  \subfloat[\label{mesh:ours_uncert1}]{
    \includegraphics[height=3.7cm]{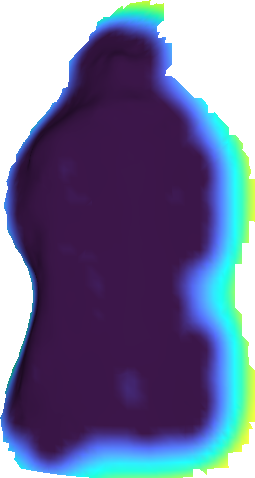}}
    \subfloat[\label{mesh:ours_uncert2}]{
    \includegraphics[height=3.7cm]{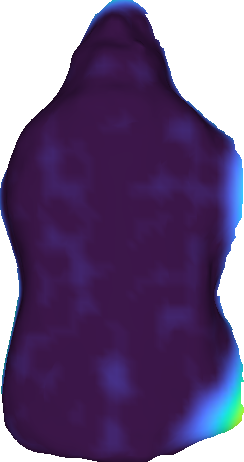}}
    \subfloat[\label{mesh:ours_uncert3}]{
    \includegraphics[height=3.7cm]{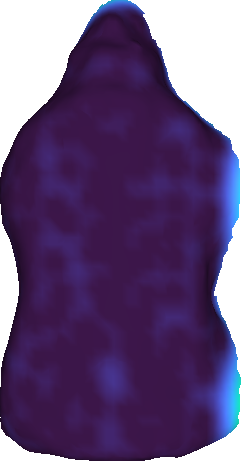}}
    \subfloat[\label{mesh:ours_uncert4}]{
    \includegraphics[height=3.7cm]{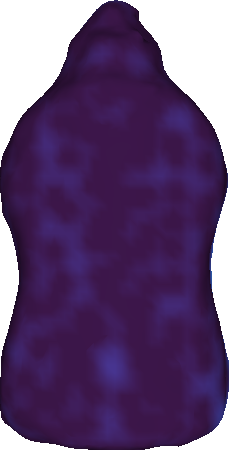}}
  \caption{Surface reconstruction results for the \texttt{006\_mustard\_bottle}. 
  From left to right: ground-truth mesh, DisFlow reconstruction, TSDF baseline reconstruction, 
  and from d) to g), we show the incremental mesh with uncertainty over frames (dark means low uncertainty). Our method produces sharper geometry and provides interpretable confidence cues.}
  \label{fig:mesh_results}
    \vspace{-3ex}
\end{figure*}

We consider the \texttt{006\_mustard\_bottle}, a challenging object. Figure~\ref{fig:mesh_results} shows a qualitative comparison: the ground-truth mesh, our reconstruction with DisFlow, a TSDF baseline reconstruction with our poses as input, and our reconstruction augmented with uncertainty colouring. The TSDF fusion tends to leave small misalignments in curved regions. In contrast, DisFlow produces smoother and more geometrically faithful surfaces. The uncertainty colouring further highlights regions with limited observations, providing valuable cues for active sensing and safe manipulation. Table~\ref{tab:mesh_eval_006} reports quantitative metrics for this sequence. DisFlow achieves a lower Chamfer distance of 0.0045 compared to 0.0059 for TSDF, indicating more accurate point-to-surface alignment. The Hausdorff distance is also reduced from 0.0106 to 0.0069. In terms of F-score, our method yields substantially higher accuracy at fine thresholds, with F@1mm improving from 0.107 (TSDF) to 0.288, and F@2mm from 0.466 to 0.645. Finally, our method achieves improved normal consistency, with the mean angular deviation reduced from 11.96$^\circ$ to 10.44$^\circ$ on the accuracy side and from 14.96$^\circ$ to 10.65$^\circ$ on the completeness side. It confirms that DisFlow reconstructs surfaces not only closer in geometry but also consistent in local orientation.

Overall, the combination of quantitative metrics and visualisations demonstrates that DisFlow provides more accurate and reliable reconstructions than the TSDF baseline. More importantly, the additional uncertainty channel makes our representation directly informative for downstream robotic tasks that require awareness of reconstruction confidence.

\begin{table}[t]
\centering
\caption{Mesh evaluation on 006\_mustard\_bottle comparing DisFlow and TSDF.}
\label{tab:mesh_eval_006}
\begin{tabular}{lcccccc}
\toprule
Method & Cf & Hf & F@1mm & F@2mm & Avg. normal ($^\circ$) \\
\midrule
TSDF         & 0.0059 & 0.0106 & 0.107 & 0.466 & 11.96 / 14.96 \\
Ours & \textbf{0.0045} & \textbf{0.0069} & \textbf{0.288} & \textbf{0.645} & \textbf{10.44} / \textbf{10.65} \\
\bottomrule
\end{tabular}
\end{table}


\subsection{Qualitative Results}
\subsubsection{Rotating Objects}
\begin{figure*}[ht]
  \centering
  \begin{subfigure}{0.195\textwidth}
    \begin{tikzpicture}
      \node[anchor=south west,inner sep=0] (big) at (0,0)
        {\includegraphics[width=\linewidth]{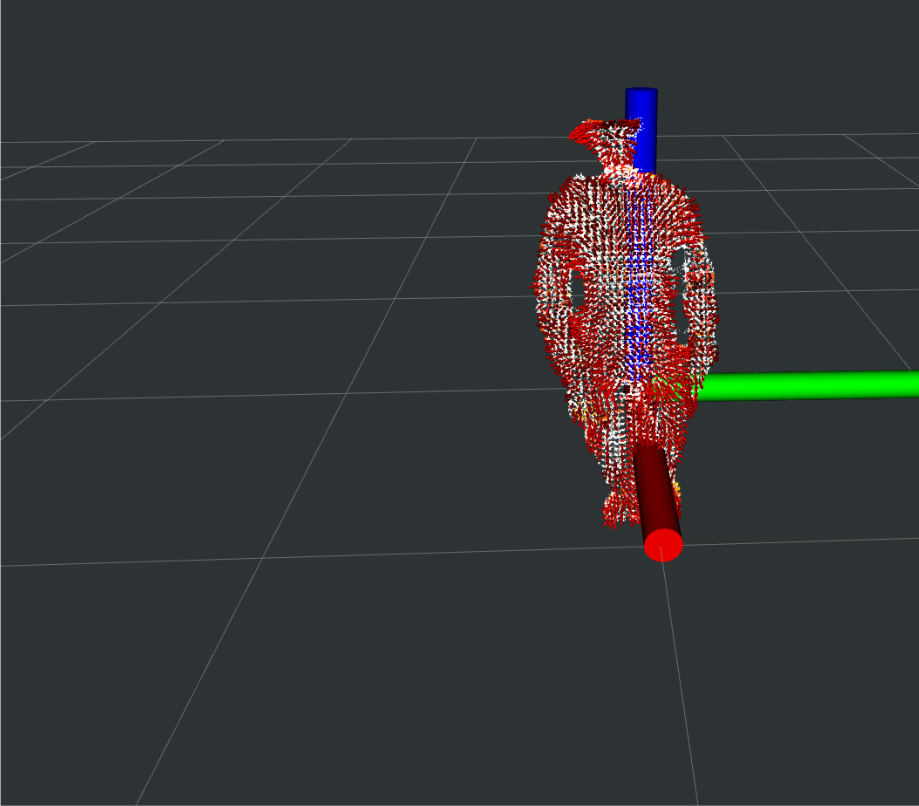}};
      \node[inset] at (big.north west){\InsetPic{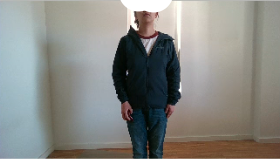}};
    \end{tikzpicture}
    \caption{front}
  \end{subfigure}\hfill
  \begin{subfigure}{0.195\textwidth}
    \begin{tikzpicture}
      \node[anchor=south west,inner sep=0] (big) at (0,0)
        {\includegraphics[width=\linewidth]{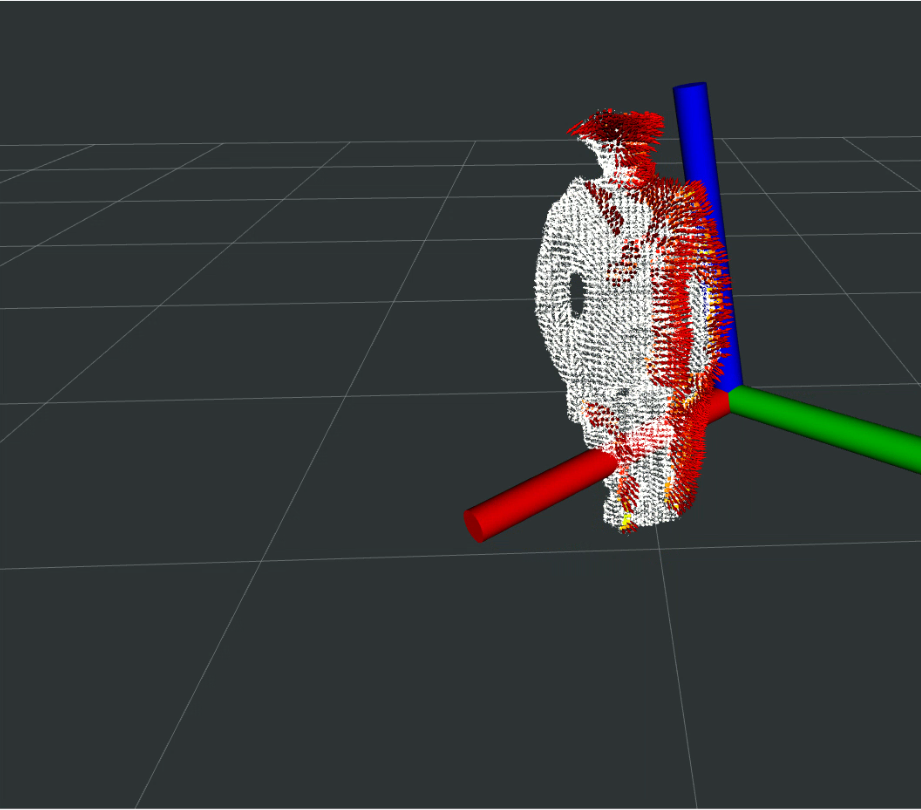}};
      \node[inset] at (big.north west){\InsetPic{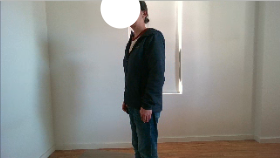}};
    \end{tikzpicture}
    \caption{left}
  \end{subfigure}\hfill
  \begin{subfigure}{0.195\textwidth}
    \begin{tikzpicture}
      \node[anchor=south west,inner sep=0] (big) at (0,0)
        {\includegraphics[width=\linewidth]{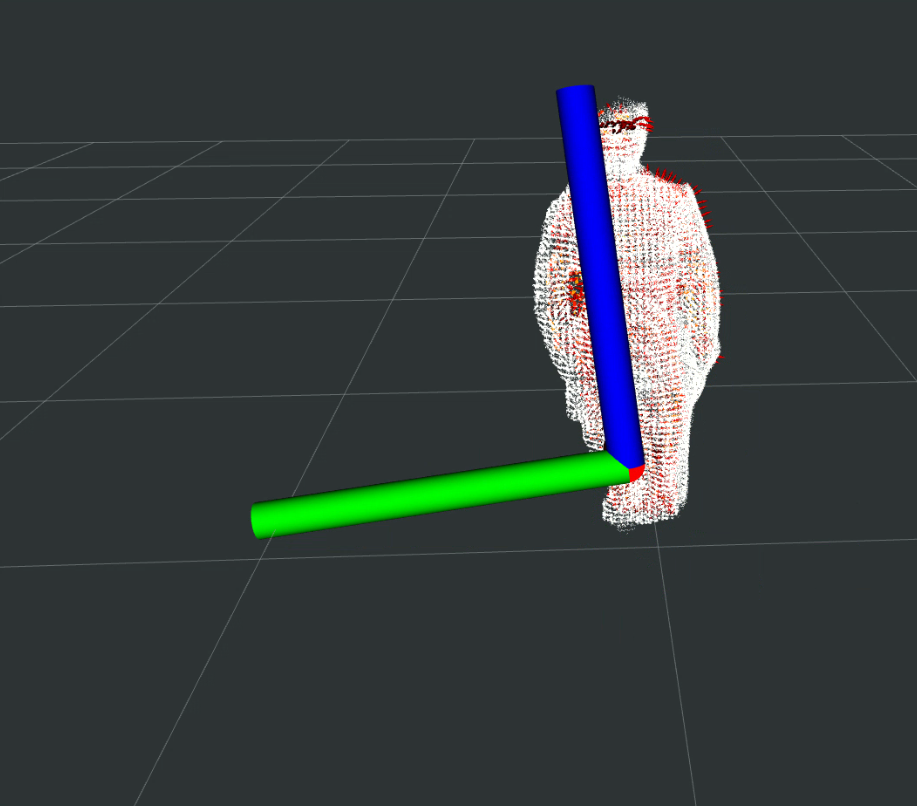}};
      \node[inset] at (big.north west){\InsetPic{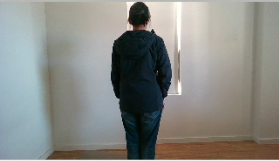}};
    \end{tikzpicture}
    \caption{back}
  \end{subfigure}\hfill
  \begin{subfigure}{0.195\textwidth}
    \begin{tikzpicture}
      \node[anchor=south west,inner sep=0] (big) at (0,0)
        {\includegraphics[width=\linewidth]{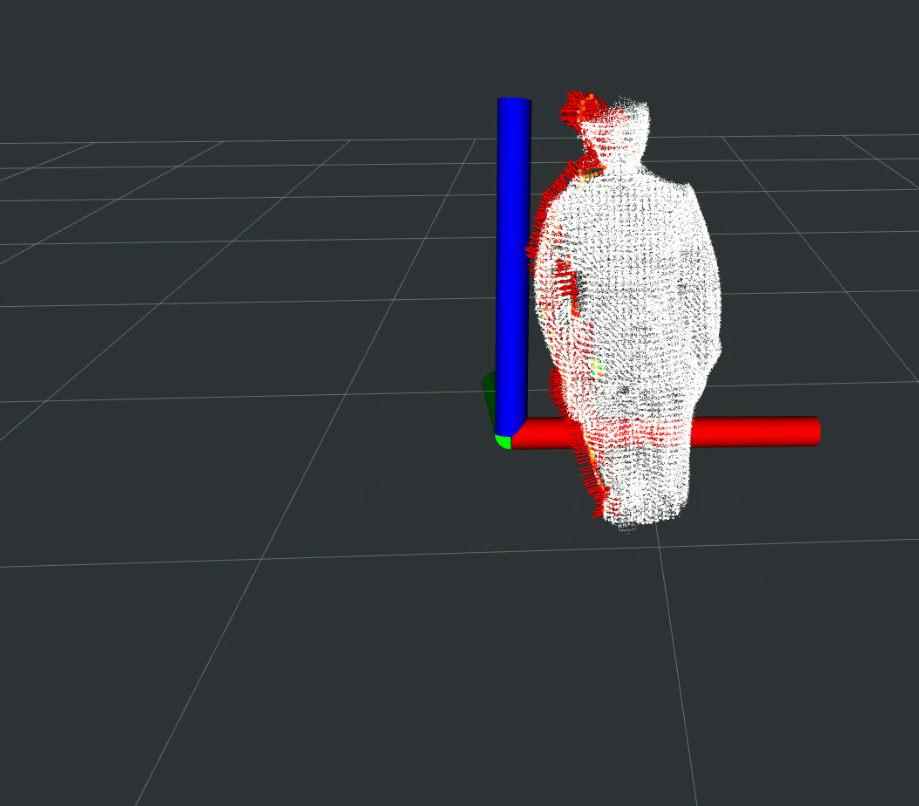}};
      \node[inset] at (big.north west){\InsetPic{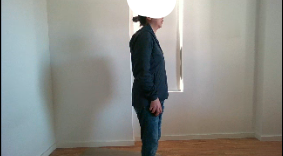}};
    \end{tikzpicture}
    \caption{right}
  \end{subfigure}\hfill
  \begin{subfigure}{0.195\textwidth}
    \begin{tikzpicture}
      \node[anchor=south west,inner sep=0] (big) at (0,0)
        {\includegraphics[width=\linewidth]{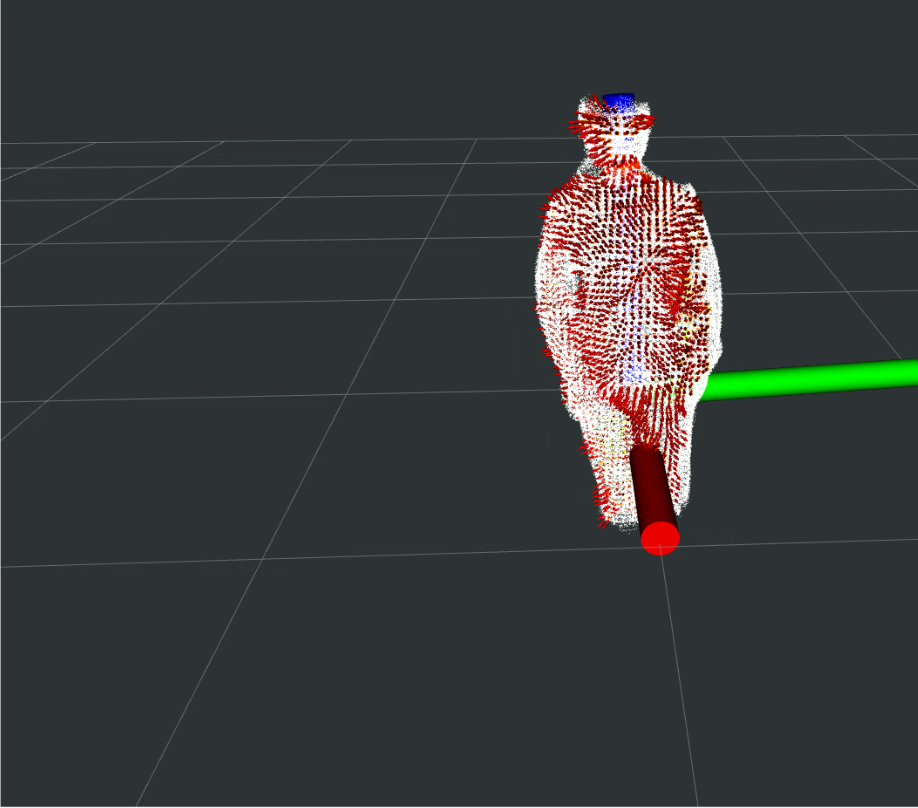}};
      \node[inset] at (big.north west){\InsetPic{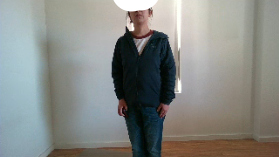}};
    \end{tikzpicture}
    \caption{back to the front}
  \end{subfigure}
  
  \caption{A full $360^{\circ}$ turn and returns to the starting pose. Each figure shows the fused dynamic human (white point cloud), the human pose with coordinate axes ($x,y,z$), and the distance flow with surface normals (red). The reconstructed human remains fixed in the object frame, and the final pose aligns with the starting pose, demonstrating consistency without drift.}
  \label{fig:human_rotation}
\end{figure*}

We first present qualitative results on a challenging real-world sequence where a human subject rotates $360^{\circ}$ in place and returns to the initial configuration. Since the subject's motion is smooth, here we can increase robustness of registration with the regularisation term detailed in Section~\ref{sec:method:practical}. Figure~\ref{fig:human_rotation} illustrates five representative snapshots spanning the full rotation. 
Each panel shows a small inset RGB image (top-left) that provides the scene context, while the main visualisation overlays our reconstructed geometry and estimated motion in the object frame. 
In all frames, the moving human is consistently fused into a dense white point cloud anchored in the object frame, remaining fixed. 
Simultaneously, the pose of the human is tracked in 6D, with the estimated coordinate axes ($x,y,z$) visualised at each instant. 
The surface geometry is further enriched with the predicted \emph{distance flow} (colour points and red arrows), highlighting both motion dynamics and geometric consistency. As the subject completes the turn and returns to the starting pose (Fig.~\ref{fig:human_rotation}e), the reconstructed point cloud remains seamless, with no visible drift or misalignment. 
This confirms that our probabilistic object-frame fusion, combined with distance-based pose tracking, produces temporally consistent reconstructions even under rotational non-rigid motions. 
Similarly in Fig.~\ref{fig:combined}, analogous results are shown for the full rotation of a toy goose.  Additional examples and full sequences are provided in the supplementary video.

\subsubsection{Flying Doughnut}
\begin{figure}[ht]
  \centering

  \begin{subfigure}{0.49\linewidth}
    \centering
    \includegraphics[height=\ImgH,keepaspectratio,trim=10mm 0 10mm 0,clip]{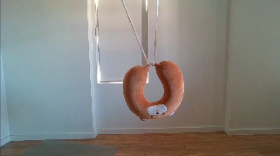}
    \caption{}
  \end{subfigure}\hfill
  \begin{subfigure}{0.49\linewidth}
    \centering
    \includegraphics[height=\ImgH,keepaspectratio,trim=80mm 0 80mm 0,clip]{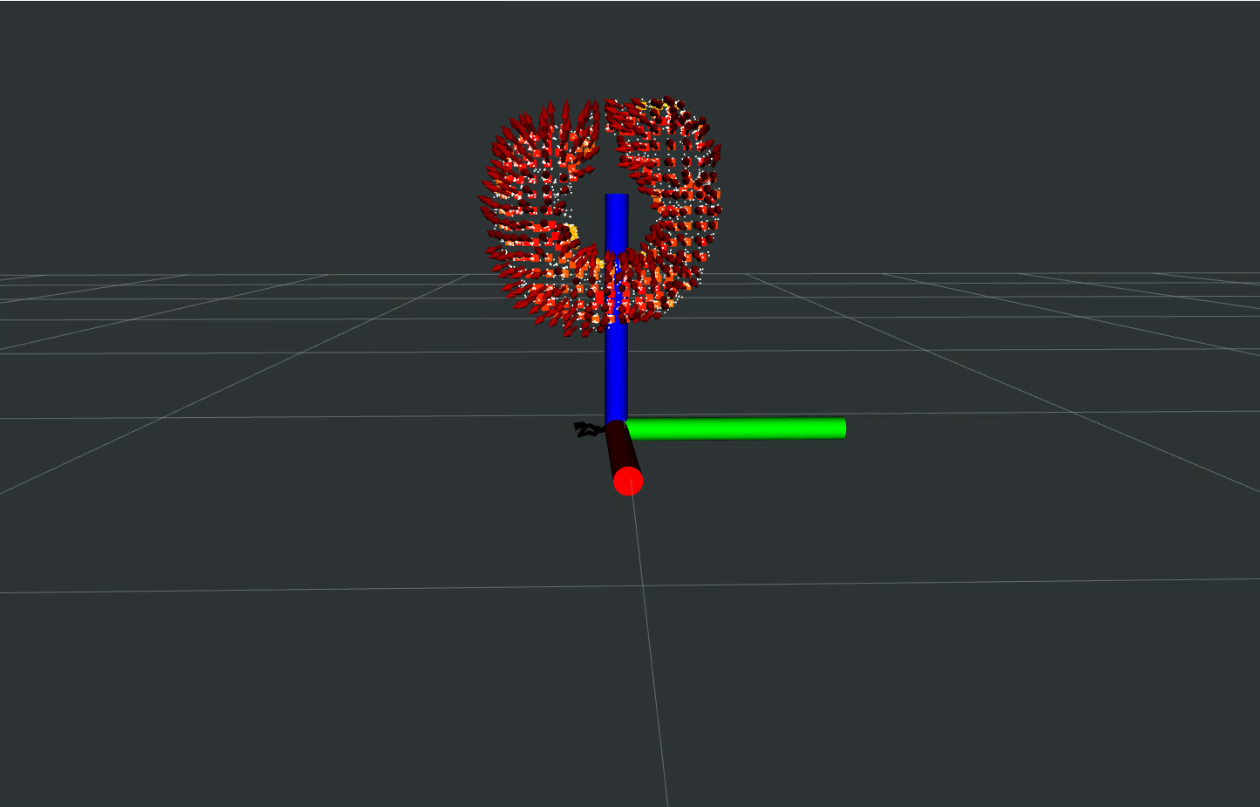}
    \caption{Starting position}
  \end{subfigure}

  \begin{subfigure}{0.49\linewidth}
    \centering
    \includegraphics[height=\ImgH,keepaspectratio,trim=10mm 0 10mm 0,clip]{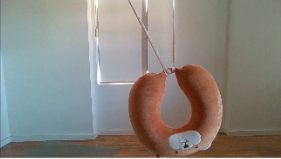}
    \caption{}
  \end{subfigure}\hfill
  \begin{subfigure}{0.49\linewidth}
    \centering
    \includegraphics[height=\ImgH,keepaspectratio,trim=80mm 0 80mm 0,clip]{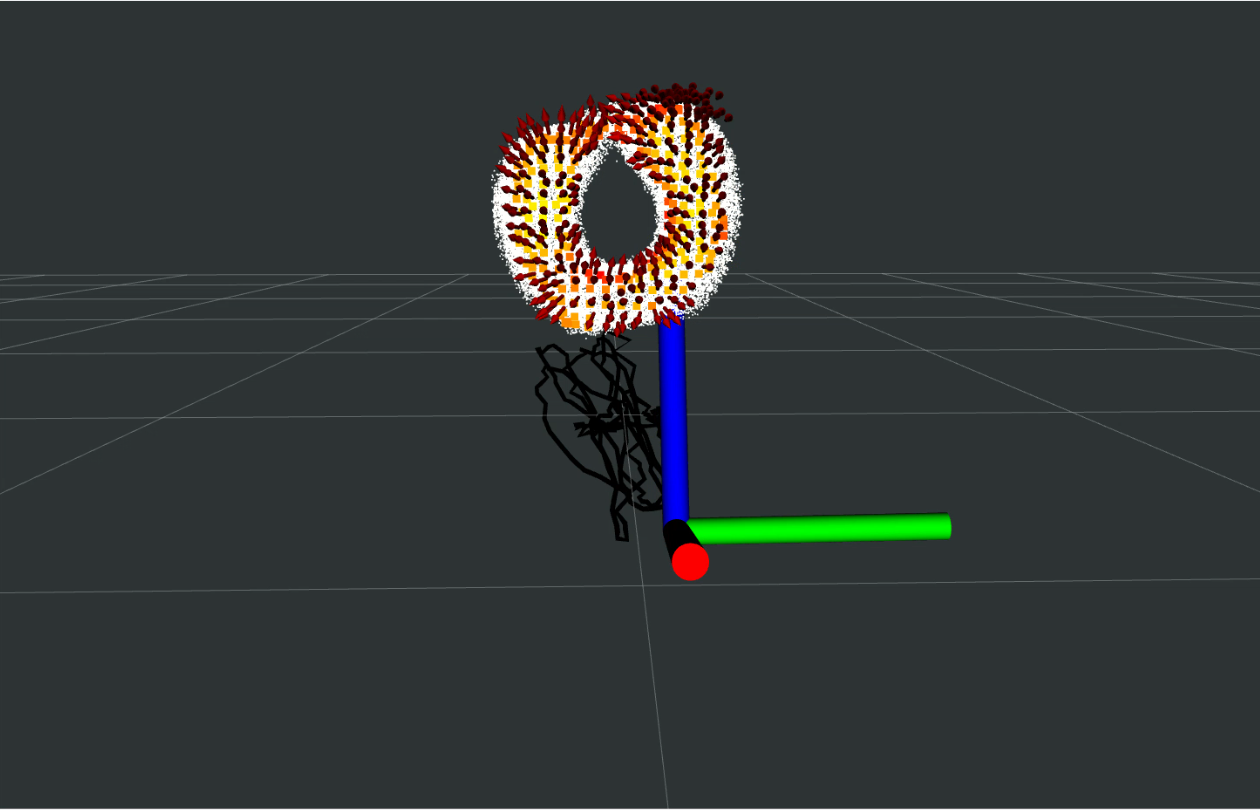}
    \caption{}
  \end{subfigure}

  \begin{subfigure}{0.49\linewidth}
    \centering
    \includegraphics[height=\ImgH,keepaspectratio,trim=10mm 0 10mm 0,clip]{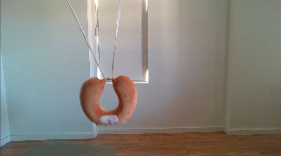}
    \caption{}
  \end{subfigure}\hfill
  \begin{subfigure}{0.49\linewidth}
    \centering
    \includegraphics[height=\ImgH,keepaspectratio,trim=80mm 0 80mm 0,clip]{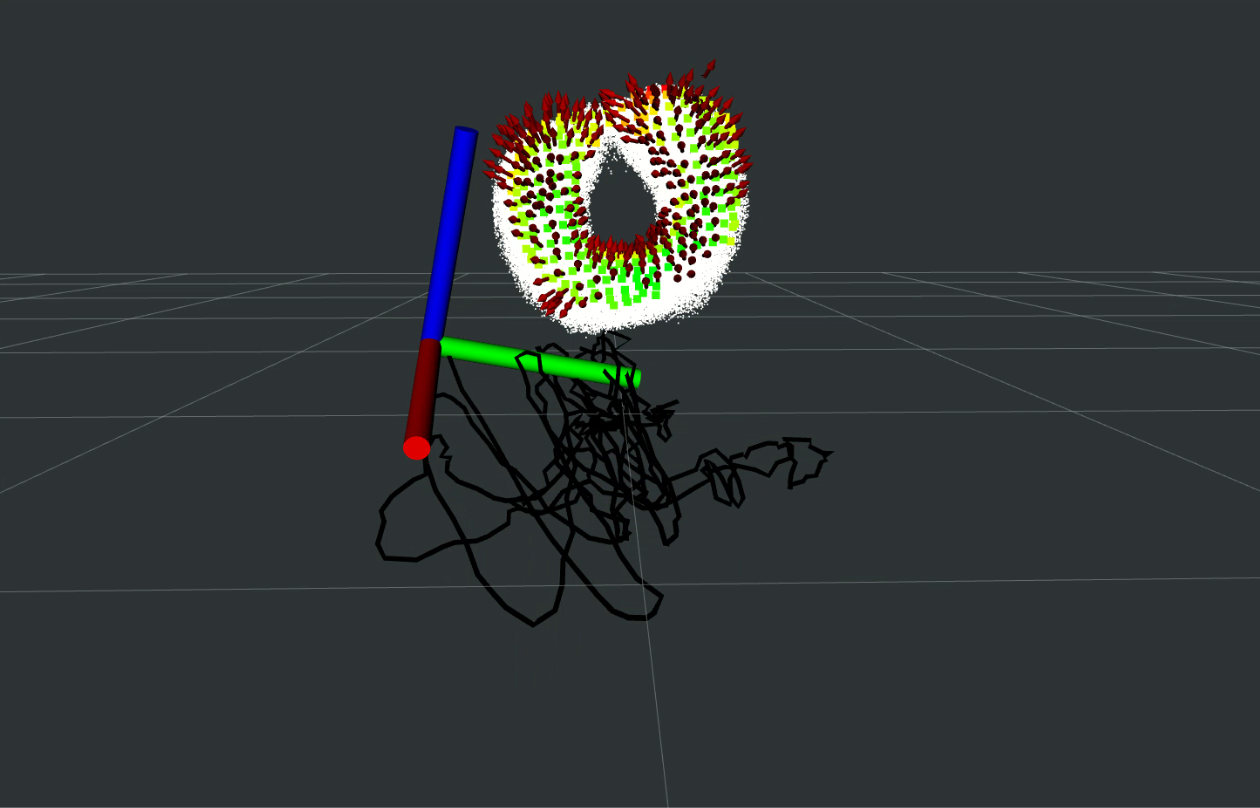}
    \caption{}
  \end{subfigure}

  \begin{subfigure}{0.49\linewidth}
    \centering
    \includegraphics[height=\ImgH,keepaspectratio,trim=10mm 0 10mm 0,clip]{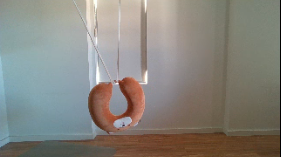}
    \caption{}
  \end{subfigure}\hfill
  \begin{subfigure}{0.49\linewidth}
    \centering
    \includegraphics[height=\ImgH,keepaspectratio,trim=80mm 0 80mm 0,clip]{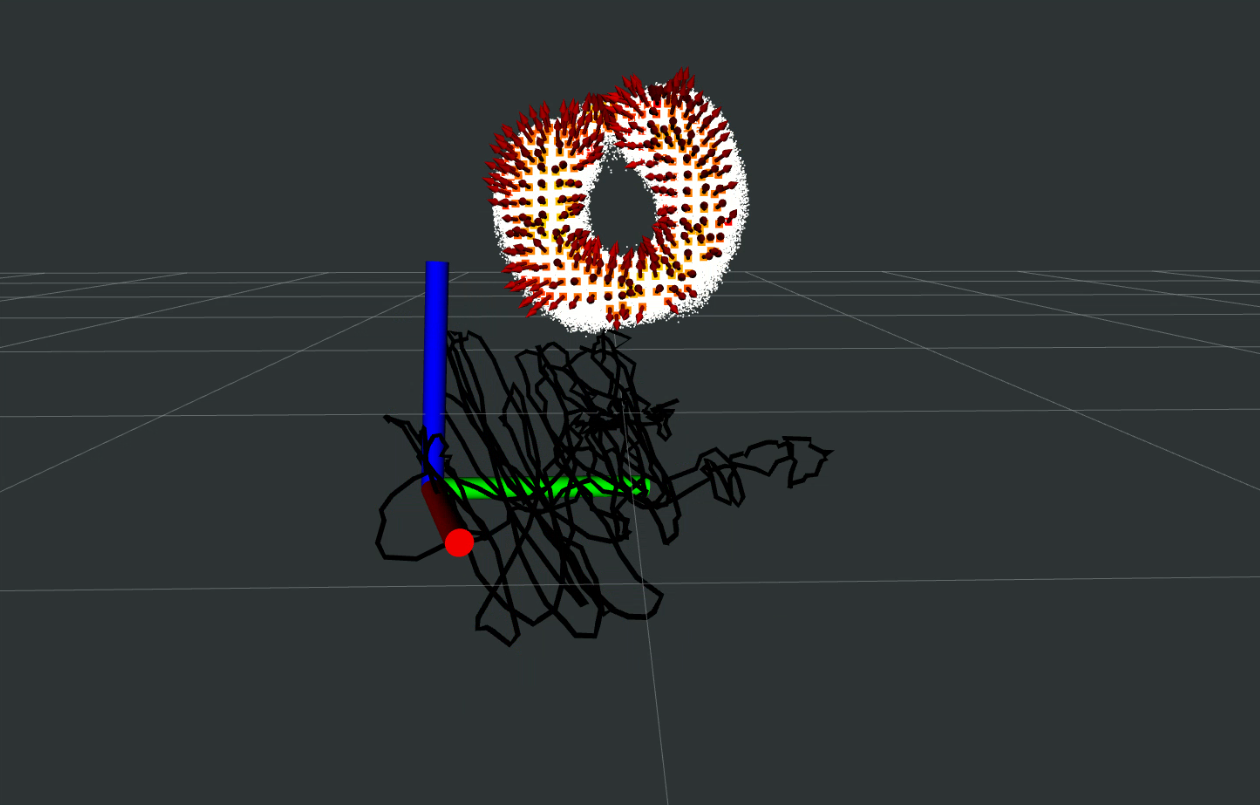}
    \caption{Back to start with an offset}
  \end{subfigure}

  \caption{Rapid motions in $\mathcal{F}_w$ frame. The motion trajectory is in black lines. The object is in the $\mathcal{F}_o$ frame (fixed).}
  \label{fig:flying_donut}
\end{figure}

We further evaluate our method on a highly dynamic scene where a flying doughnut undergoes rapid translational and rotational motions in Figure~\ref{fig:flying_donut}. 

Similar to the human rotation experiment, the object measurements are accumulated into a consistent white point cloud anchored in the object frame, remaining fixed despite drastic changes. 
The doughnut's estimated 6D poses are visualised with coordinate axes in the world frame, and its motion trajectory is additionally highlighted by black lines connecting successive poses. 
This trajectory clearly illustrates the fast and non-smooth dynamics of the object, which are faithfully captured by our framework. In addition, the predicted distance flow is shown by colour points and red arrows. 
Qualitative results are demonstrated in our video.

\section{Conclusion and Future Work}

We have introduced \textit{DisFlow}, a unified framework for real-time object pose and velocity tracking together with dense probabilistic surface reconstruction. By leveraging Gaussian Process Implicit Surfaces in the object frame, our method couples pose estimation and fusion through the notion of scene flow, enabling temporally consistent geometry, accurate motion estimates, and principled uncertainty. DisFlow achieves state-of-the-art accuracy for both 6D pose and velocity tracking. Beyond tracking, our approach delivers high-quality reconstructions. Crucially, the probabilistic formulation provides uncertainty estimates that can be visualised on reconstructed meshes and directly integrated into downstream planning, manipulation, and interaction tasks.



While the results are promising, our method is not without limitations. For objects with surface structure, DisFlow provides reliable constraints that lead to accurate pose tracking and reconstruction. However, when objects are nearly flat, possess minimal structure or high symmetry (e.g., bottles or cans), the constraints become weaker and the optimisation may converge to ambiguous solutions. 
One possible solution would be to incorporate RGB image features or appearance-based correspondences into the registration process. 
However, in the present work, we intentionally focus on the geometry-only formulation, and we leave the integration of visual features to future work.

\section{Acknowledgement}
This work was carried out while the first author was with the University of Technology Sydney.

\bibliographystyle{IEEEtran}
\bibliography{reference}

\end{document}